\documentclass[DIV=calc, paper=a4, fontsize=11pt, twocolumn]{scrartcl}	 

\usepackage{lipsum} 
\usepackage[english]{babel} 
\usepackage[protrusion=true,expansion=true]{microtype} 
\usepackage{amsmath,amsfonts,amsthm} 
\usepackage[svgnames]{xcolor} 
\usepackage[hang, small,labelfont=bf,up,textfont=it,up]{caption} 
\usepackage{booktabs} 
\usepackage{fix-cm}	 

\usepackage{times}
\usepackage{epsfig}
\usepackage{graphicx}
\usepackage{amsmath}
\usepackage{amssymb}
\usepackage{array}
\usepackage{caption}
\usepackage{xcolor}
\usepackage{authblk}
\usepackage{flushend}

\usepackage[margin=0.8in,letterpaper]{geometry} 
\usepackage{cite} 
\usepackage[final]{hyperref} 
\hypersetup{
	colorlinks=true,       
	linkcolor=blue,        
	citecolor=blue,        
	filecolor=magenta,     
	urlcolor=blue         
}

\usepackage{sectsty} 
\allsectionsfont{\usefont{OT1}{phv}{b}{n}} 

\usepackage{fancyhdr} 
\usepackage{lastpage} 

\newcommand\eg{\emph{e.g}}
\newcommand\etal{\emph{et.al}}

\usepackage{lettrine} 

\usepackage{titling} 

\newcommand{\HorRule}{\color{DarkGoldenrod} \rule{\linewidth}{1pt}} 

\pretitle{{ \fontsize{20}{5.5cm}\selectfont {Technical Report}} \vspace{-45pt} \begin{flushleft} \HorRule \fontsize{50}{50} \usefont{OT1}{phv}{b}{n} \color{DarkRed} \selectfont} 

\title{Recurrent Convolutional Neural Network Regression for Continuous Pain Intensity Estimation in Video} 

\posttitle{\par\end{flushleft}\vskip 0.5em} 

\preauthor{\begin{flushleft}\large \lineskip 0.5em \usefont{OT1}{phv}{b}{sl} \color{DarkRed}} 

\author{Jing Zhou$^{1,2}$, Xiaopeng Hong$^{1}$\thanks{\textit{Corresponding author}}, Fei Su$^{2}$, Guoying Zhao$^{1}$}

\postauthor{\footnotesize \usefont{OT1}{phv}{m}{sl} \color{purple} 

$^1$ITEE, University of Oulu \\
$^2$Beijing University of Posts and Telecommunications \\ 

{\color{gray}{\textit {Emails: 1522670460@qq.com, xhong@ee.oulu.fi, sufei@bupt.edu.cn, gyzhao@ee.oulu.fi}}}
\par\end{flushleft}\HorRule} 
\date{} 

\begin{document}

\maketitle
\thispagestyle{empty}

\begin{abstract}
Automatic pain intensity estimation possesses a significant position in healthcare and medical field. Traditional static methods prefer to extract features from frames separately in a video, which would result in unstable changes and peaks among adjacent frames. To overcome this problem, we propose a real-time regression framework based on the recurrent convolutional neural network for automatic frame-level pain intensity estimation. Given vector sequences of AAM-warped facial images, we used a sliding-window strategy to obtain fixed-length input samples for the recurrent network. We then carefully design the architecture of the recurrent network to output continuous-valued pain intensity. The proposed end-to-end pain intensity regression framework can predict the pain intensity of each frame by considering a sufficiently large historical frames while limiting the scale of the parameters within the model. Our method achieves promising results regarding both accuracy and running speed on the published UNBC-McMaster Shoulder Pain Expression Archive Database.

\end{abstract}

\section{Introduction}

Measuring or monitoring pain intensity is crucial in pain medication, treatment or diagnosis to individuals who are unable to communicate verbally, such as newborns and patients in intensive care units. Normally, pain intensity measurements are conducted via self-report or checked by medical staffs (\eg, nurse or physician). But these measurements may cause unreliability or a large workload of hospitals. Thus, a reliable automatic pain intensity estimation model provides a more economical option to measure pain intensity of different subjects.

In the past decade, a plenty of approaches have been proposed for automatic pain intensity estimation. Table~\ref{tab:survey} provides a brief summary of typical approaches. Early researches tend to focus on estimating whether the subject is painful or not, and thus, conduct pain intensity estimation as a classification problem ~\cite{ref5}, ~\cite{ref6}, ~\cite{ref7}, ~\cite{lucey2009automatically11}, ~\cite{lucey2012painful12}, ~\cite{khan2013pain13}, ~\cite{pedersen2015learning16}, ~\cite{rathee2015novel17}.

\begin{table*}[!htbp]
\begin{center}
\begin{tabular}{l|c|c|c|c}
\hline
Feature descriptors & Pain levels & Measures  & Classifier & Cross Validation \\
\hline\hline
C-APP + S-PTS ~\cite{ref5} & C-2 & OPI, PSPI & SVM & Leave One Subject Out\\
PTS + APP ~\cite{lucey2009automatically11} & C-2 & PSPI & SVM & Leave One Subject Out\\
PTS, APP ~\cite{ref7} & C-2 & PSPI & SVM & Leave One Subject Out\\
SAPP +SPTS + CAPP ~\cite{lucey2009automatically11} & C-2 & PSPI & SVM+LLR & Leave One Subject Out\\
AAM ~\cite{lucey2012painful12} & C-2 & OPI, PSPI & SVM & Leave One Subject Out\\
PLBP, PHOG ~\cite{khan2013pain13} & C-2 & PSPI & SVM & 10-fold\\
Auto Encoder ~\cite{pedersen2015learning16} & C-2 & PSPI & SVM & Leave One Subject Out\\
TPS ~\cite{rathee2015novel17} & C-2	& PSPI & DML + SVM	& Leave One Subject Out \\
Canny Edge ~\cite{ref8} & C-2/C-8 &	OPI, PSPI &	TBM	& 3-fold\\

LBP ~\cite{chen2012person9} & C-2 &	PSPI	& Transfer Learning &	Leave One Subject Out\\
PCA ~\cite{adibuzzaman2015assessment10} & C-3	& VAS	& SVM, Angular Distance &	10-fold\\
DCT + LBP ~\cite{kaltwang2012continuous33} & R	& PSPI	& RVR & Leave One Subject Out \\
Hess + Grad + AAM ~\cite{ref4} & R	& PSPI	& SVM	& Leave One Subject Out\\
2Standmap~\cite{hong2015capturing34} & R & PSPI &	RVR	& Leave One Subject Out\\
\hline
\end{tabular}
\end{center}
\caption{A brief summary of the recent methods proposed for pain detection and pain intensity estimation.\\
1st column - Feature descriptors: S-PTS: Similarity Normalized Shape, S-APP: Normalized Appearance, C-APP: Canonical Appearance, PTS: Normalized Shape, APP: Appearance, DCT: Discrete Cosine Transform, LBP: Local Binary Pattern, AAM: Active Appearance Model, PLBP: Pyramid LBP, PHOG: Pyramid Histogram of Orientation Gradients, TPS: Thin Plate Spline, PCA: Principal Component Analysis, Hess: Hessian based histograms, Grad: Gradient-based histograms; 2nd column - Pain levels: C: classification, R-n: n-level regression; 3rd column - Measures of pain intensity: OPI: Observer Pain Intensity, PSPI: Prkachin and Solomon Pain Intensity, VAS: Visual Analog Scale; 4th column - Classifier: SVM: Support Vector Machine, RVR: Relevance Vector Regression, NN: Nearest Neighbor, LLR: Linear Logistic Regression, TBM: Transferable Belief Model; and last column - Manner of Cross Validation.}
\label{tab:survey}
\end{table*}

More recently, an increasing number of researchers realize that simply judging whether it is painful or not for a whole sequence is too rough for fine-grained pain intensity estimation in practice. Therefore, they start to study frame-level pain intensity estimation and regard it as a regression problem.

One crucial issue here is to provide enough data where each frame is well labeled under a standard scientific measure to facilitate related researches. In 2008, Prkachin and Solomon~\cite{ref1} proposed a measure of pain intensity termed by Prkachin and Solomon Pain Intensity (PSPI) based on Facial Action Coding System (FACS)~\cite{ref2},~\cite{prkachin1992consistency21}. PSPI is defined as a function of the intensity of six pain related Facial Action Units (AUs), which describe a set of facial configurations related to pain such as nose wrinkling and cheek-raising. 
By using PSPI as the frame-level intensity measure, a few recent works have been proposed for pain intensity regression. Kaltwang \etal. ~\cite{kaltwang2012continuous33} compared three approaches by using the locations of 66 facial landmark points, DCT, and LBP, as well as the combinations among them. Florea \etal. used the histogram of topographical features and SVM, achieving a great result of average mean squared error (MSE) ~\cite{ref4}. In ~\cite{hong2015capturing34}, Hong \etal. applied a second-order standardized moment average pooling (2Standmap) method which beats all approaches that only rely on a single descriptor.

However, traditional static features like LBP and DCT, which are extracted from separate frames, have inevitable limitations in describing relevant dynamic information required by pain intensity estimation. For example, subjects tend to close eyes when they are suffering pain, but traditional features and static methods cannot differentiate between normal eye blink or eye closure that related to pain from independent frames. It thus results in unstable changes and peaks of the estimation among adjacent frames.

To overcome this problem, we attempt to encode the video not only from the separate frames but also among adjacent frames. In this paper, we propose a regression framework based on Recurrent Convolutional Neural Network (RCNN) for automatic frame-level pain intensity estimation. In the first step, we used Active Appearance Model (AAM) to track faces and warped all facial images of different poses. In the second step, given the vector sequences of the warped facial images, we used a sliding-window strategy to achieve fixed-length input samples of the recurrent network from the video sequence. Finally, we carefully design the architecture of the recurrent convolutional neural network for continuous-valued pain intensity. The proposed end-to-end pain intensity regression framework can predict the pain intensity of each frame by considering a sufficiently large historical frames while limiting the scale of the parameters within the model.

The main contribution of this work is that we propose an RCNN based framework to estimate pain intensity automatically. According to the best knowledge of the authors, it is the first time that the recurrent (convolutional) neural network is applied to the task of pain intensity estimation. Correspondingly, the RCNN is used as an end-to-end regressor, which outputs continuous scores rather than discrete labels as in the problem of classification.

The proposed regression network is evaluated on the published UNBC-McMaster Shoulder Pain Expression Archive Database, where our method gets promising results with a real-time testing speed.

The remaining content of this paper is organized as follows. Section~\ref{sec:related_work} briefly introduces the background of RCNN. Section~\ref{sec:rnn} details the proposed framework. Quantitative experimental results are provided in Section~\ref{sec:exp}. Section~\ref{sec:con} concludes the paper.

\section{Recurrent Convolutional Neural Network}
\label{sec:related_work}

\begin{figure*}
\begin{center}
\includegraphics[width=0.98\linewidth]{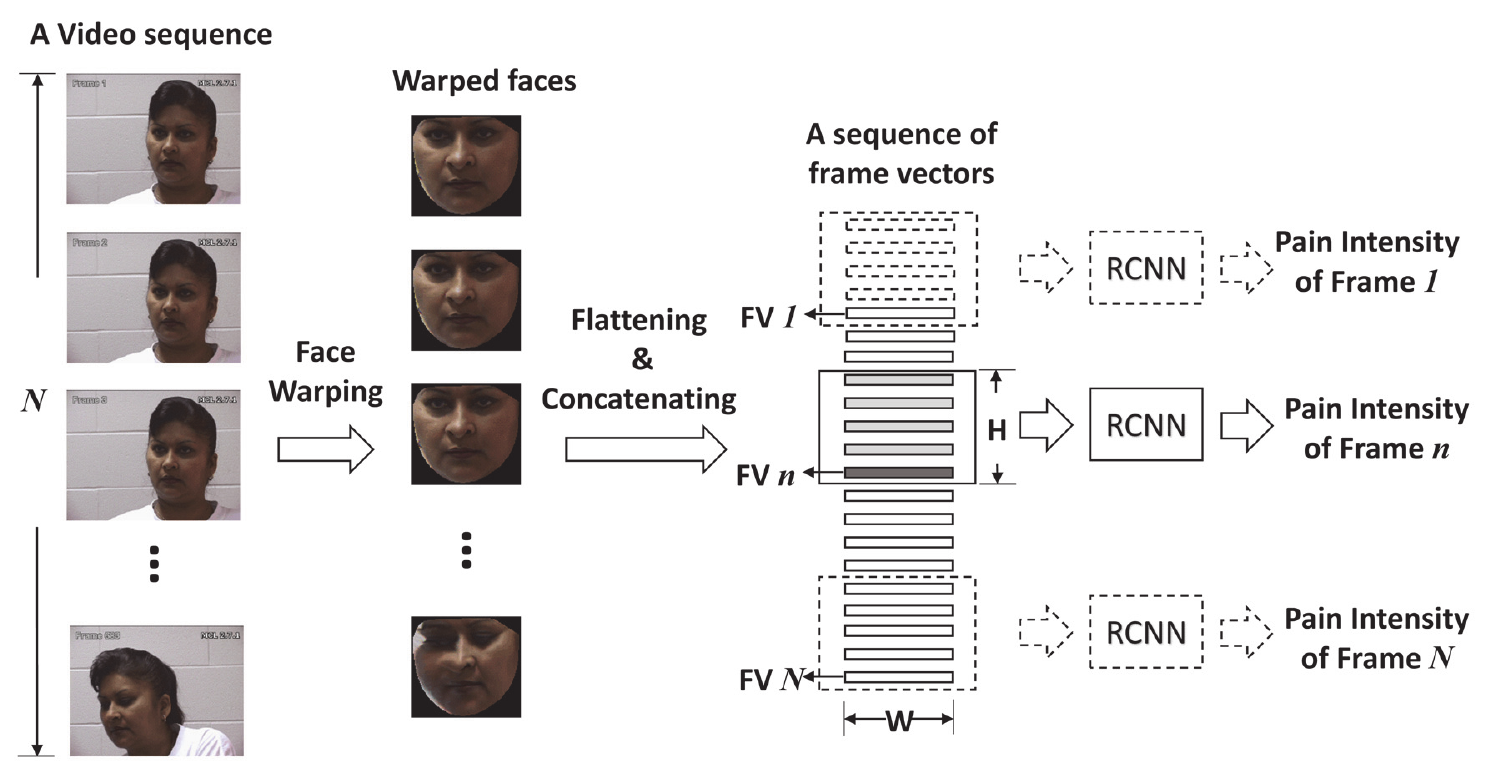}
\end{center}
   \caption{The framework of the proposed pain intensity estimation approach.}
\label{fig:framework}
\end{figure*}


In the past few years, Convolutional neural network (CNN) has made a great success in various computer vision tasks, such as image classification ~\cite{szegedy2015going22}, object detection ~\cite{girshick2014rich23}, and tracking ~\cite{wang2015transferring24}. CNN has been characterized by local connections, weight sharing, and local pooling, which largely attribute to excellent performances.
Recurrent neural network (RNN) has a long history in the artificial neural network community~\cite{carpenter1987massively25},~\cite{elman1990finding26},~\cite{fernandez1990nonlinear27}, the most successful applications refer to sequential tasks such as ~\cite{graves2009novel28},~\cite{graves2013speech29}. RNN has been characterized by connecting hidden layers of the current time step and several previous time steps. Because RNN reserves the temporal information in sequences, it achieves a great performance in sequential tasks.
Combining the advantages of CNN and RNN, different structures of networks were proposed to fuse convolutional layers and recurrent layers to capture relevant contextual information from raw pixels in static images. In 2014, Pinheiro and Collobert ~\cite{Pinheiro2014Recurrent30} used extra recurrent connections from the top layer to the bottom layer of a CNN for scene labeling. In~\cite{liang2015recurrent19}, Liang and Hu proposed an RCNN for object recognition by using recurrent connections within the same layer. Their models are different from our proposed RCNN regression, which lies in two folds: first, the RCNNs in ~\cite{Pinheiro2014Recurrent30} and \cite{liang2015recurrent19} are applied to the tasks based on static images while here we used RCNN for modeling the temporal information in videos; secondly, their RCNNs are used as classifiers by using softmax function as the activation function of the fully connected layer, ours is used as a regressor for estimate pain intensity. The architecture of normal RCNN will be explained in Section~\ref{sec:rnn}.

\section{Frame-by-Frame Regression Network}
\label{sec:rnn}


\subsection{The Framework}

The key problems of pain intensity estimation can be summarized as four blocks. Firstly, each incoming facial frame ($n$) of the testing video sequence should be aligned and warped to the same frontal pose. Secondly, in order to keep spatial and temporal information at the same time, we need to convert each warped face into a (3-channel (RGB)) frame vector (FV$n$). Thirdly, because of the fixed height (H) of the input that our RCNN requires, we applied a sliding window to achieve testing samples. When testing the frame n, the testing sample contains several continuous adjacent frame vectors before the frame $n$ (padding zeros if $n<$H). Finally, we fed the samples to a trained RCNN, and the network will output the PSPI predictions frame by frame. The whole framework is shown as Fig.~\ref{fig:framework}.

As for the training process, we used a random strategy to achieve the training samples of fixed length. Similarly, we converted all frame images into frame vectors. All converted frame vector sequences will be immersed in a training pool; then the network uses windows to select randomly a subset of the training data to conduct one training iteration. The length of every training sample (H) indicates the number of continuous frames that the recurrent network will use at one time. Then, these training samples will be fed into the RCNN regression structure to start learning. 

\subsection{Preprocessing}

\begin{figure}
\begin{center}
\includegraphics[width=0.98\linewidth]{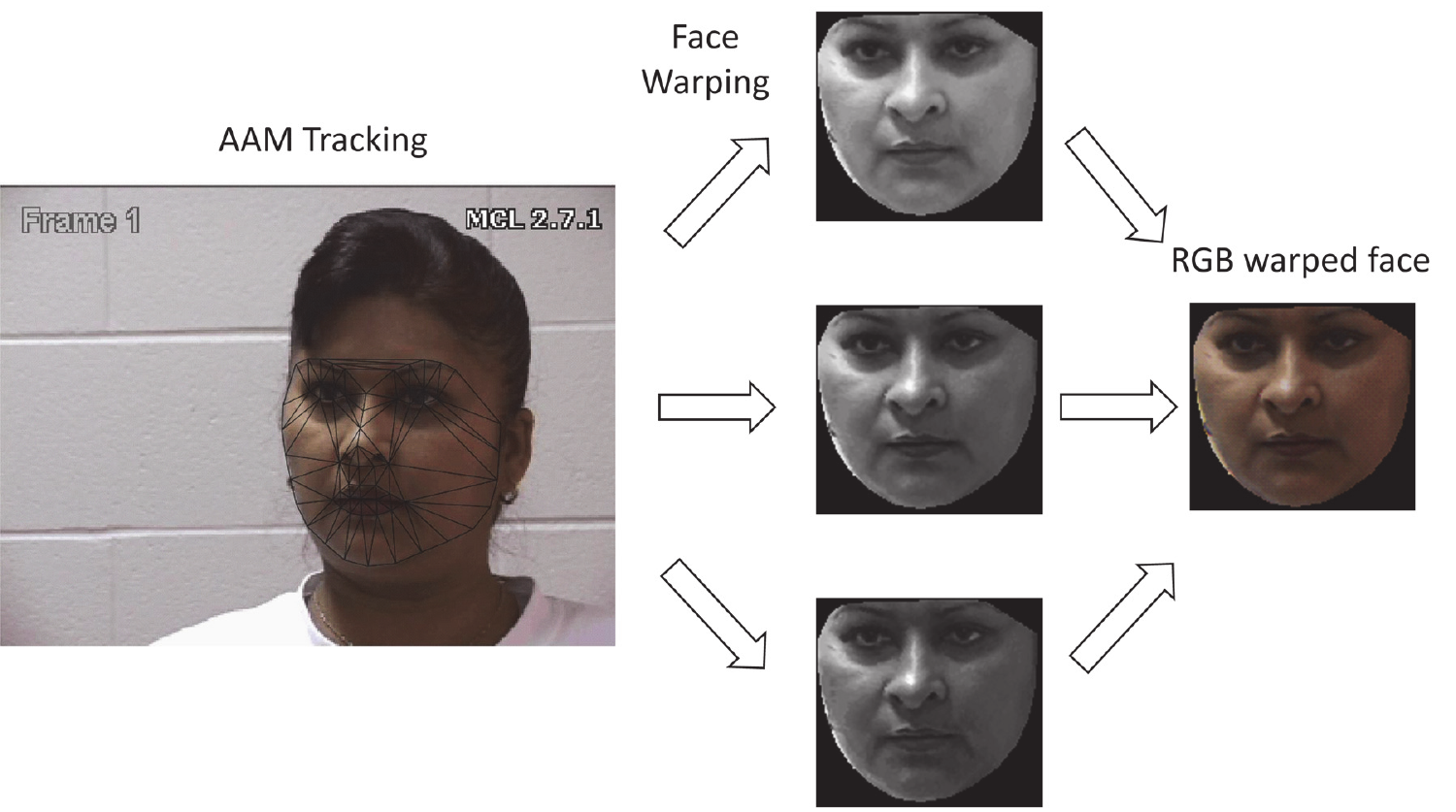}
\end{center}
   \caption{AAM tracking; R, G, B warped faces; and RGB warped face.}
\label{fig:aam}
\end{figure}

A pain intensity estimation algorithm should be both robust to face pose and the identity of the subject (not subject dependent). To achieve invariance to different face poses, we exploited an Active Appearance Model (AAM) to warp all facial images of different poses into the same frontal pose. AAM tracks the face and extracts visual features, finding the key points on faces, such as eyebrows, and the outline of faces. These AAM landmark points constitute many non-overlapping triangles, which can warp and align different faces into the same 2D triangulated mesh after some linear shape variation ~\cite{cootes2001active31}, ~\cite{tzimiropoulos2013optimization32}. 
In the process of face aligning and warping, we used the same facial triangulated mesh for all subjects. We warped every facial image in RGB channels separately, then combined all channels back to get the final RGB warped faces (see Fig.~\ref{fig:aam}).

The input samples of our RCNN structure should be no more than two dimensions, but to reserve the temporal information among frames and the spatial pixel information of warped facial images at the same time, we considered some different ways to convert each frame into a 1D vector, such as flattening or extracting feature vectors. Finally, it turned out that flattening is an effective way though it may lose some structural information of the images. After flattening, we concatenated all 1D flattened warped facial images in frame order to achieve frame vector sequences. 

\subsection{Architecture of RCNN}

\begin{figure}
\begin{center}
\includegraphics[width=0.98\linewidth]{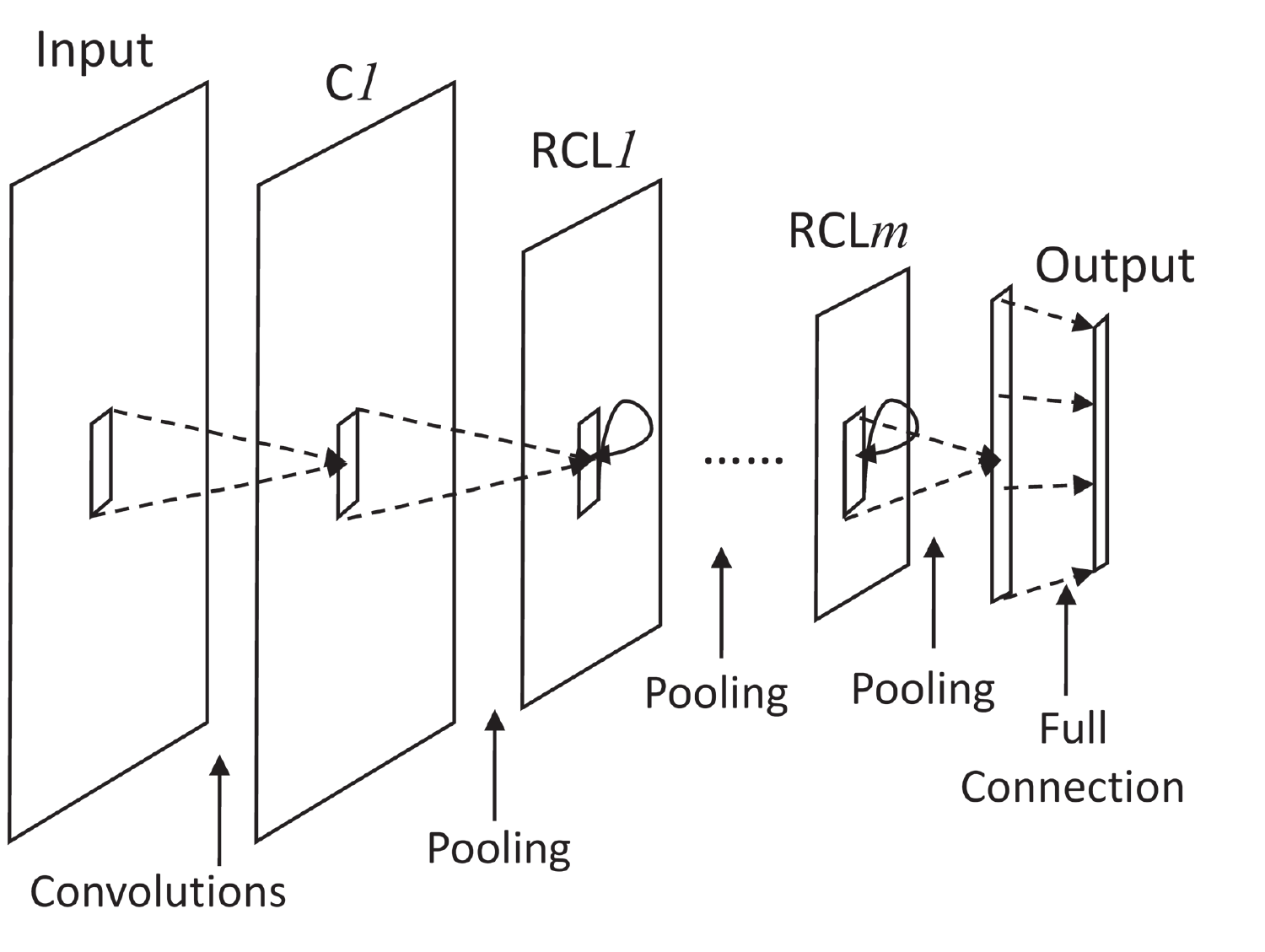}
\end{center}
   \caption{The overall architecture of RCNN.}
\label{fig:rnn_archi}
\end{figure}

The basic idea of RCNN is to add recurrent connections within every convolutional layer of the feed-forward CNN ~\cite{liang2015recurrent19}. The overall architecture of RCNN is shown in Fig.~\ref{fig:rnn_archi}. The first layer (C\textit{1}) is the standard feed-forward convolutional layer without recurrent connections. Following (C\textit{1}), there are several recurrent convolutional layers (RCL\textit{1}$\sim$RCL$m$), with a max pooling layer between every two RCLs. Normally, the final output layer is a softmax layer in the tasks of classification.

\begin{figure}
\begin{center}
\includegraphics[width=0.98\linewidth]{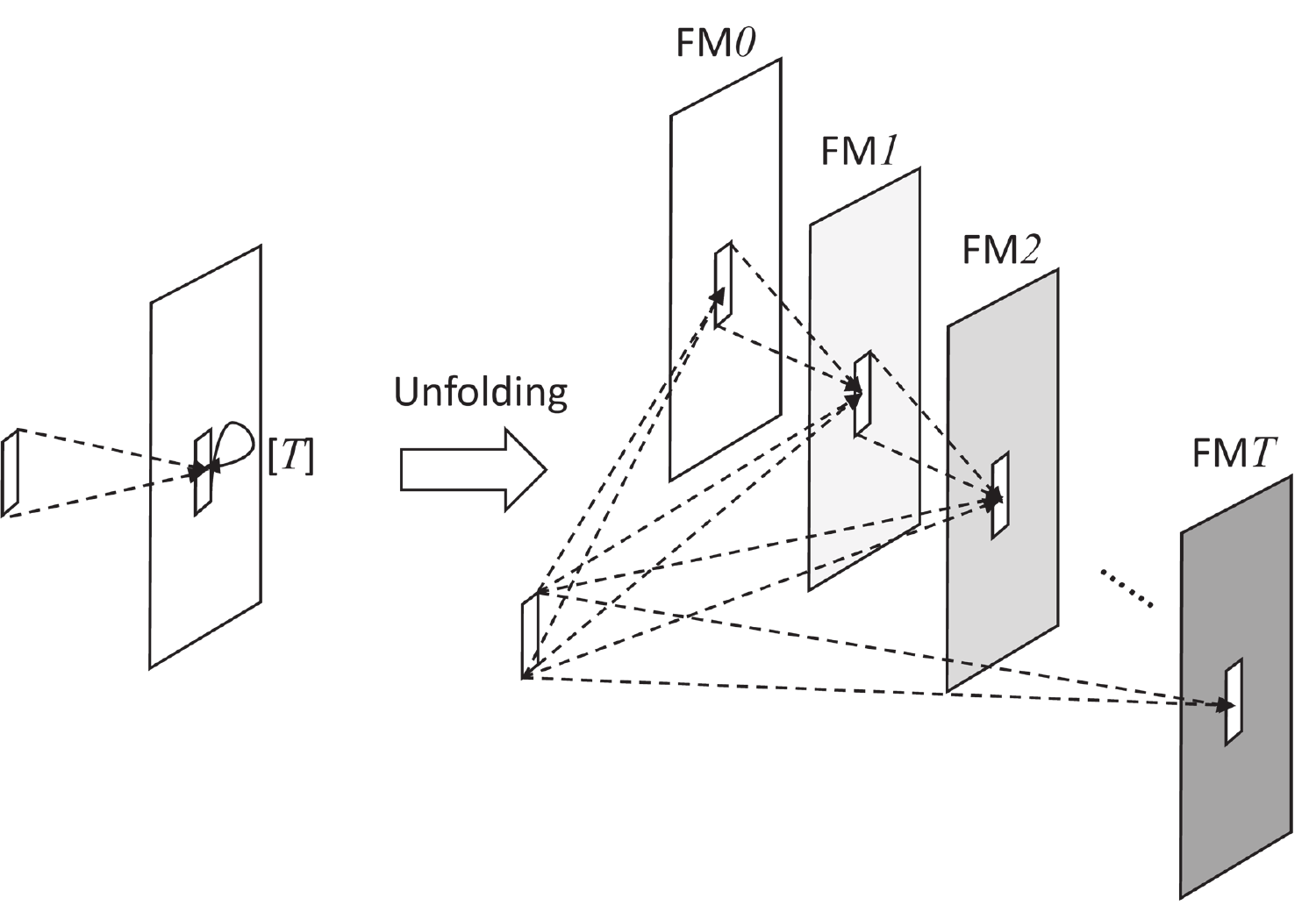}
\end{center}
   \caption{Unfolding an RCL.}
\label{fig:rnn_unfolding}
\end{figure}

Each RCL is constituted by several iterative convolutions, sharing weights in hidden layers among $T+1$ time steps. If unfolding an RCL, the layer can be seen as a feed-forward subnetwork with the depth of $T+1$ (see Fig.~\ref{fig:rnn_unfolding}). The difference between an RCL and a $(T+1)$-layer CNN is that the inputs of RCL are all the values of time steps from 0 to $T$, but the inputs of CNN are the values of one fixed time step. Thus, unfolding an RCNN through time steps in RCLs can result in an arbitrarily deep network with a fixed number of parameters. 

The overall depth of the model is crucial for obtaining good o the performance~\cite{zeiler2014visualizing20}. The existence of deeper layers or longer paths among layers in a network makes it possible for the network to learn highly complex features. On the contrary, shorter paths may help gradient backpropagation during training. RCNN is actually a CNN with flexible paths between the input layer to the output layer, which expands the depth of the network but also facilitate the learning ~\cite{liang2015recurrent19}. In the structure of an RCL, there are several paths from the first feature map of convolution (FM\textit{0}) to the last feature map (FM$T$). In Fig.\ref{fig:rnn_unfolding}, the darker the feature map is, the deeper the path is. Attribute to the iteration in an RCL, the length of path ranges from 1 to $T+1$, including the first path of the convolutional layer. 
In our framework, we used four ($m=4$) RCLs in the whole RCNN architecture. Therefore, the length of the iterative path will range from 6 to $4(T+1)+2$ including the first path of C\textit{1} and the last path of the output layer. The length of recurrent time steps ($T$) was empirically set as 3. 

The following subsection will introduce how the output layer is modified for continuous-valued predictions. And more detailed implementation setup of our network will be described in Section~\ref{sec:exp}.

\subsection{Continuous Predictions}

The recurrent convolutional neural network (RCNN) are usually used to solve classification problems such as image classification~\cite{liang2015recurrent19} and scene labeling~\cite{Pinheiro2014Recurrent30}.
Correspondingly, to assign feature vectors to one of the \emph{C} categories, the final output layer is a \emph{softmax} layer whose output is given by:
\begin{equation}
\label{eq:softmax}
\hat{y}_i=\frac{\exp \left ( \mathbf{w}^{T}_i \mathbf{x}\right)}{\sum_{i'} \mathbf{w}^{T}_{i'}\mathbf{x} },
\end{equation}

\noindent where $ \hat{y}_i $ is the predicted probability belonging to the $i$th category, for $i = 1, 2,..., C$, and $\mathbf{x}$ is the feature vector generated by the global max pooling before the output layer. The training process is performed by minimizing the cross-entropy loss function as:
\begin{equation}
\label{eq:cross_entropy}
L=-\frac{1}{N}\sum_{i=1}^{N}\left [ y_i\log\hat{y}_i+\left ( 1-y_i \right )\log\left ( 1-\hat{y}_i \right ) \right ].
\end{equation}

As for the estimation of pain intensity, the network should allow continuous-valued predictions. A linear function is therefore simply used as the activation function in the output layer of the network:

\begin{equation}
\label{eq:linear}
\hat{y}=\mathbf{w}^{T} \mathbf{x},
\end{equation}

\noindent where $ \hat{y} $ is the continuous predicted value of the network, and $\mathbf{x}$ is the feature vector. Correspondingly, the loss function is modified to the mean squared error function as:
\begin{equation}
\label{eq:mseloss}
L=\frac{1}{N}\sum_{i=1}^{N}\left ( \hat{y}_i-y_i\right )^{2} ,
\end{equation}

\noindent rather than the cross-entropy function in Eq.~\ref{eq:cross_entropy}. With it, the output becomes continuous so that it turns a regressor. Training is performed by minimizing the MSE function using the back-propagation through time (BPTT) algorithm ~\cite{werbos1990backpropagation39}. This is equivalent to using the standard BP algorithm on the time-unfolded network. The final gradient of a shared weight is the sum of its gradients over all time steps.

\section{Experiments}
\label{sec:exp}

\subsection{Pain Intensity Dataset}

\begin{figure}
\begin{center}
\includegraphics[width=0.98\linewidth]{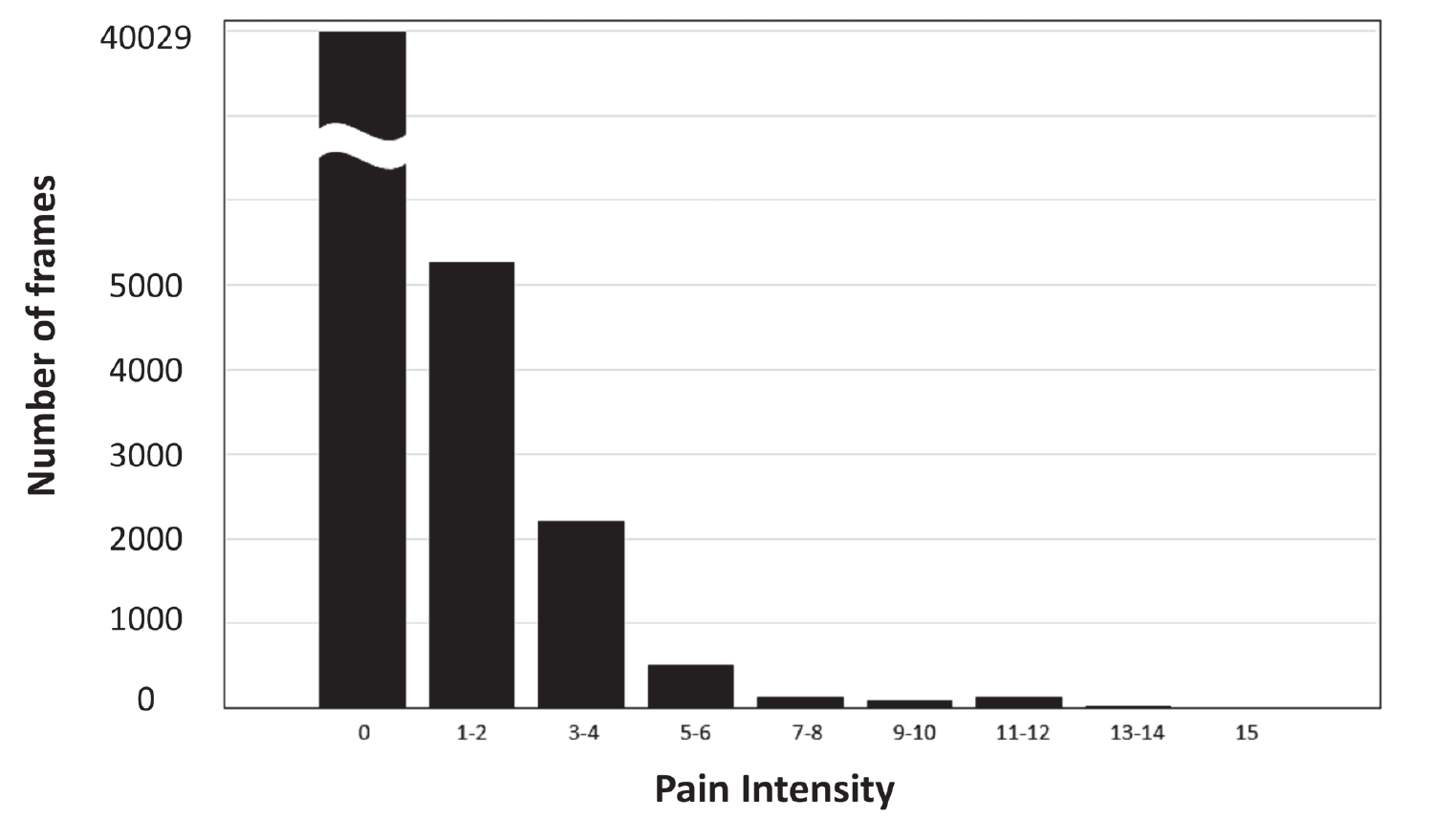}
\end{center}
   \caption{Frame distribution of the PSPI (0-15).}
\label{fig:frame_stat}
\end{figure}

Recently, researchers at the McMaster University and University of Northern British Columbia (UNBC) published a shoulder pain expression archive database ~\cite{ref3}. This database is the most common database to be used to assess pain detection or pain intensity estimation methods.
The database captured face videos of subjects (66 females and 63 males) when they were performing a series of active and passive range-of-motion tests to their affected and unaffected limbs on two separate occasions. Out of which videos of active tests are publicly available for research purposes. In this database, each video was coded by FACS in frame level. Observer and self-report measurements in sequence level were also taken. The PSPI score was computed to quantify pain intensity in 16 discrete levels (0-15) based on AUs ~\cite{ref2},~\cite{prkachin1992consistency21}.
In this paper, we used the videos of active tests to perform pain intensity estimation experiments, with the 16-level PSPI as the ground-truth. Active tests include 200 sequences of 25 subjects, with totally 48,398 frames of 320$\times$240 pixels. We noticed that the frame distribution of the PSPI is quite unbalanced as shown in Fig.~\ref{fig:frame_stat}.

To solve the unbalanced training samples of 16 levels, we designed a weighted strategy to keep training samples of all labels balanced to some extent. The network selects a subset of the training samples randomly to conduct one training iteration. The subset contains samples of all PSPI levels, and the percentage of samples corresponding to each PSPI level is weighted manually.
\subsection{Measurement}

In our experiments, we conducted a leave-one-subject-out strategy which leads to 25-fold cross-validation to assess our method. We left all sequences of one chosen subject as the testing set and the rest sequences of 24 subjects as the training set at the same time. The average Mean Squared Error (MSE) and Pearson Product-moment Correlation Coefficient (PCC) were calculated by:
\begin{equation}
\label{eq:mse}
\textit{MSE}=\frac{1}{N}\sum_{i=1}^{N}\left ( \hat{y}_i-y_i\right )^{2} ,
\end{equation}

\begin{equation}
\label{eq:pcc}
\textit{PCC}=\frac{\sum _{i=1}^{N}\left ( \hat{y}_i-\overline{\hat{y}_i} \right )\left (y_i-\overline{y} \right )}{\sqrt{\sum _{i=1}^{N}\left ( \hat{y}_i-\overline{\hat{y}_i} \right )^2}\sqrt{\sum _{i=1}^{N}\left (y_i-\overline{y} \right )^2}},
\end{equation}

\noindent where $N$ is the total number of frames of testing sequences. $y_i$ and $\hat{y}_i$ are the ground-truth and the pain intensity estimation of the $i$ frame, respectively. $\overline{\hat{y}_i}$ and $\overline{y}$ are the sample mean of $ \left \{ y_1,...,y_N \right \} $ and $ \left \{ \hat{y}_1,...,\hat{y}_N \right \} $.
\subsection{Implementation Details}

As is described in Section~\ref{sec:rnn}, we got 3-channel (RGB) frame vector sequences (H$\times$W) as the input of the network. The choice of H is strongly related to the time cycle of the pain occurrence. In our experiments, H and W were empirically set as 30 and 713, respectively. In each RCL, we used one convolutional layer first (functioning as a feed-forward layer), then connected three iterations ($T=3$ in Fig.\ref{fig:rnn_unfolding}) following the feed-forward layer. In the fully connected layer, we used a linear function as the activation to conduct the regression task and the MSE function as the loss measurement. A summary of the main network configurations is shown in Table~\ref{tab:net_config}.

\begin{table}[!htbp]
\begin{center}
\begin{tabular}{l|l}
\hline
Layer type & Configurations\\
\hline\hline
Input	& W(713)$ \times$ H(30) $\times 3$ \\
 & RGB vector sequence \\
 \hline
Convolution l	& maps:256, $k:3\times3, s:1$\\
\hline
Max pooling 1	& $p:4\times1, s:4\times1$\\
\hline
RCL 2	& feed-forward map:256 $k:1\times1, s:1$\\
 & 3 iteration maps:256 $k:3\times3, s:1$\\
 \hline
Max pooling 2 &	$p:4\times1, s:4\times1$\\
\hline
RCL 3	& feed-forward map:256 $k:1\times1, s:1$\\
 & 3 iteration maps:256 $k:3\times3, s:1$\\
 \hline
Max pooling 3 &	$p:4\times4, s:4\times4$\\
\hline
RCL 4	&  feed-forward map:256 $k:1\times1, s:1$\\
 & 3 iteration maps:256 $k:3\times3, s:1$\\
 \hline
Max pooling 4 & 	$p:2\times2, s:2\times2$\\
\hline
RCL 5	& feed-forward map:256 $k:1\times1, s:1$\\
 & 3 iteration maps:256 $k:3\times3, s:1$\\
 \hline
Max pooling 5	&  $p:1\times1, s:1\times1$\\
\hline
Output	& H(30) predictions\\
\hline
\end{tabular}
\end{center}
\caption{A summary of the main network configurations. $k, s$ and $p$ stand for the kernel size, stride and pooling size in the related layer, respectively.}
\label{tab:net_config}
\end{table}

The initial learning rate was set heuristically and annealed according to a schedule pre-determined on the cross-validation set. When the accuracy improved so slowly, we decreased the learning rate to its 1/10. Annealing was used three times through a whole training process so that the final learning rate was 1/1000 of the initial value. The momentum was fixed at 0.9. Weight decay decreased overfitting as well as dropout. Moreover, we used a batch normalization technique~\cite{ioffe2015batch35} following the first convolutional layer and every feed-forward layer in RCLs to accelerate the training process.
We implemented the network within the Theano 0.8 ~\cite{bastien2012theano37}, ~\cite{bergstra2010theano38} framework. Our experiments were carried out on a workstation with two 2.30GHz Intel(R) Xeon(R) E5-2650 v3 CPU, 320GB RAM, and an NVIDIA(R) Tesla K80 GPU to run our experiments. The average testing time is 25 frame per second.

\subsection{Experimental Results}

In our experiments, we compared our method with the state-of-the-arts on the UNBC-McMaster Shoulder Pain Expression Archive Database as shown in Table~\ref{tab:results}. 

\begin{table}[!htbp]
\begin{center}
\begin{tabular}{l|c|c}
\hline
Methods &	MSE &	PCC\\
\hline\hline
PTS ~\cite{kaltwang2012continuous33} & 2.59 &	0.36\\
DC
~\cite{kaltwang2012continuous33} & 1.71 &	0.55\\
LBP ~\cite{kaltwang2012continuous33} & 1.81 &	0.48\\
(DCT+LBP)/RVR ~\cite{kaltwang2012continuous33} & 1.39	& 0.59\\
2Standmap ~\cite{hong2015capturing34} & 1.42	& 0.55\\
Hessian Histograms ~\cite{ref4}&3.76&	0.25\\
Gradient Histograms ~\cite{ref4}&4.76&	0.34\\
Hess+Grad ~\cite{ref4}&3.35&	0.41\\
VGG-face CNN SVR&	1.70&	0.43\\
RCNN regression &	\textbf{1.54}&	\textbf{0.65}\\
\hline
\end{tabular}
\end{center}
\caption{Comparison of the proposed approach with other approaches in the literature.}
\label{tab:results}
\end{table}

\begin{figure*}[!thbp]
\begin{center}
\includegraphics[height=0.98\linewidth,angle =-90]{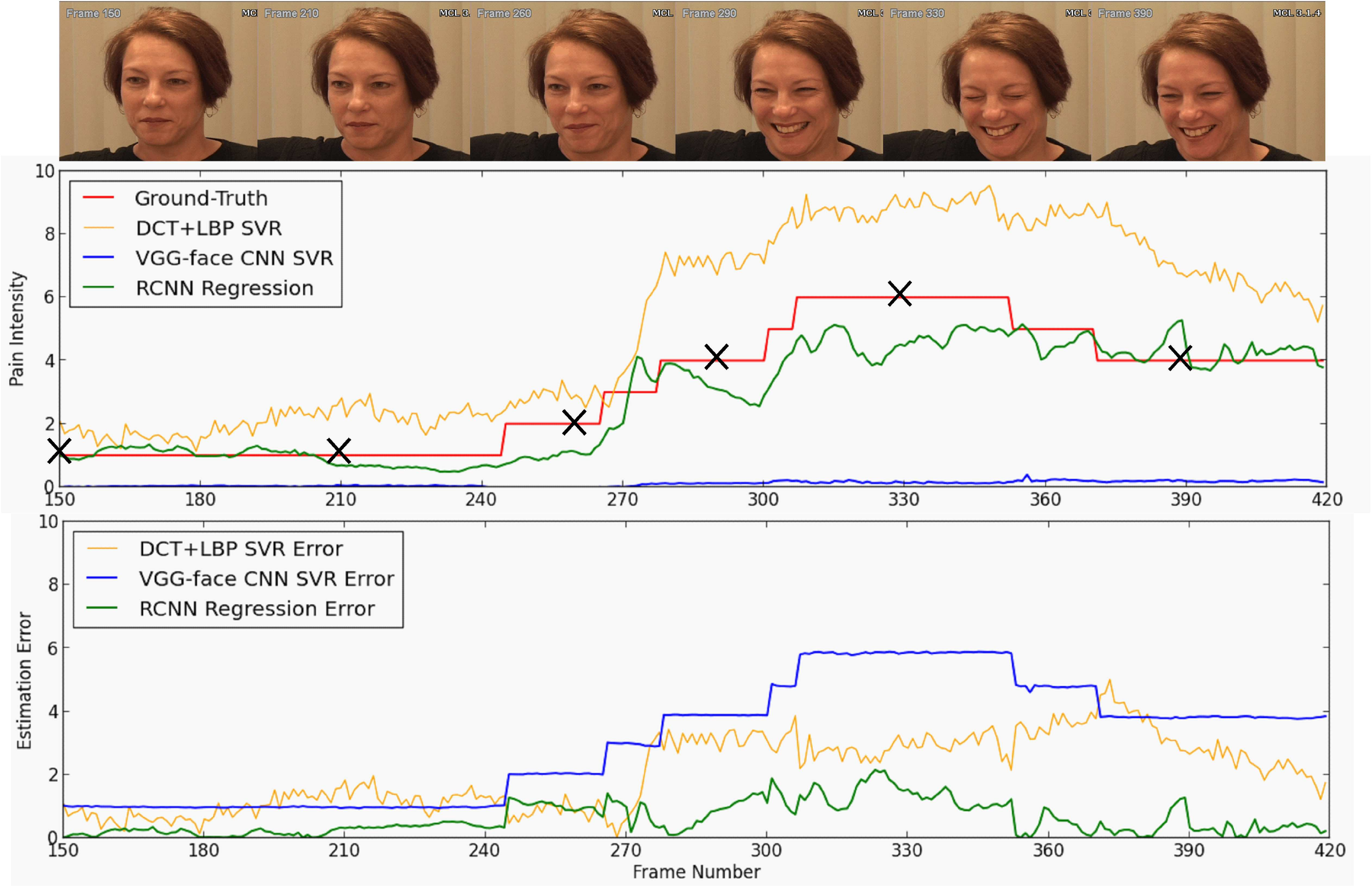}
\end{center}
   \caption{An example sequence of pain intensity and estimation error for DCT+LBP SVR, VGG-face CNN SVR, and RCNN Regression. (N.B. The "X" on the Ground-Truth correspond to the frames of number 150, 210, 260, 290, 330, and 390).}
\label{fig:pain_example}
\end{figure*}

Single features, mean feature fusion, and RVR feature fusion were proposed in ~\cite{kaltwang2012continuous33}, which includes the combinations of DCT and LBP. The mean feature fusion method calculates the weighted mean of the responses of the regression function based on one single descriptor directly, and the RVR feature fusion method using Relevance Vector Regression. ~\cite{ref4} extracts Hessian based histograms, gradient based histograms and AAM landmarks as features and uses SVM as the classifier, getting the best average MSE among all methods. ~\cite{hong2015capturing34} applies a second-order standardized moment average pooling (2Standmap) method which beats all approaches that only rely on a single descriptor. Additionally, we also used a method by extracting CNN features (VGG-face CNN SVR) as a baseline method of neural networks. We fed all warped facial images into the VGG-face CNN~\cite{parkhi2015deep36}, then we delivered the VGG-face descriptors to linear SVR~\cite{scikit-learn40}.

As for our proposed method, we used regression RCNN to conduct pain intensity estimation. We got promising results of the average MSE and PCC of 1.54 and 0.65, respectively. It indicates that our method is effective. Regarding the computational speed, our method was able to process 25 frames per second on our workstation (two 2.30GHz Intel(R) Xeon(R) E5-2650 v3 CPU, 320GB RAM, and an NVIDIA(R) Tesla K80 GPU). Therefore, our method is testing efficient for real time application.
Fig.~\ref{fig:pain_example} shows an example pain intensity estimation sequence (frame 150 to 420) of one subject using: DCT+LBP SVR, VGG-face CNN SVR, and our proposed RCNN regression. These methods got the MSE of 3.83, 10.06, 1.12 and the PCC of 0.90, 0.55, 0.89 respectively in this sequence. Compared to the other two methods, RCNN regression has a smoother approximation and smaller estimation error. Besides, from the frame 270 on, the subject appears to close her eyes. Normally, eye closure relates to pain to some extent. Using traditional structural features (\eg, LBP and DCT) and static methods (trained per frame) cannot differentiate eye blink (short time) and eye closure (long time), so the model tends to result in that all eye-closed images are strongly related to pain as it has learned in the training stage. It is the exact reason that the estimation of pain intensity by using DCT+LBP SVR keeps a continuous high level after the frame 270. However, our proposed regression RCNN is a dynamic method that predicts one frame by using several adjacent frames, which keeps the estimation line stable, smooth, and closed to the ground-truth.

\section{Conclusion}
\label{sec:con}

In this paper, we propose an automatic frame-by-frame pain intensity estimation framework in video based on a regression recurrent convolutional neural network. By leveraging the RCNN, firstly, the proposed framework predicts the pain intensity of each frame by considering a sufficiently large historical frames while limiting the scale of the parameters within the model; secondly, the framework encodes the spatial information, without losing temporal information of videos. To achieve continuous pain intensity estimation frame by frame, we modify the loss and the activation functions in the last fully connected layer of normal RCNN so that it has an output of continuous values.
The proposed method is evaluated the UNBC-McMaster Shoulder Pain Expression Archive Database. The comparisons with state-of-the-art methods are promising. We also show that the output of the proposed method turned out stable, smooth, and also can avoid unstable jumps or peaks among frames which are inevitable via static methods. Last but not least, our method is computationally efficient for real-time applications. Future work may study accelerating the training section of the RCNN for pain intensity estimation.

\section*{Acknowledgment}

This work is sponsored by the Academy of Finland, Infotech Oulu and Tekes Fidipro Program. Moreover, Xiaopeng Hong is partly supported by the Natural Science Foundation of China under the contract No. 61572205. Finally, we appreciate Mr. Ming Liang for sharing the codes of the recurrent convolutional neural network.

{\small
\bibliography{rcnn_pain_ref}
}


\end{document}